\documentclass{article} 
\usepackage{iclr2023_conference,times}
\usepackage{graphicx} 
\usepackage{listings}
\usepackage{xcolor}

\usepackage{amsmath,amsfonts,bm}

\def\eqref#1{equation~\ref{#1}}

\def\1{\bm{1}}

\DeclareMathAlphabet{\mathsfit}{\encodingdefault}{\sfdefault}{m}{sl}
\SetMathAlphabet{\mathsfit}{bold}{\encodingdefault}{\sfdefault}{bx}{n}

\definecolor{keycolor}{rgb}{0.95, 0.95, 0.95} 
\definecolor{stringcolor}{rgb}{0.2, 0.6, 0.8} 
\definecolor{numbercolor}{rgb}{0.8, 0.4, 0.4} 

\lstdefinelanguage{json}{
    basicstyle=\ttfamily,
    showstringspaces=false,
    breaklines=true,
    frame=single,
    backgroundcolor=\color{keycolor},
    literate=
     *{0}{{{\color{numbercolor}0}}}{1}
      {1}{{{\color{numbercolor}1}}}{1}
      {2}{{{\color{numbercolor}2}}}{1}
      {3}{{{\color{numbercolor}3}}}{1}
      {4}{{{\color{numbercolor}4}}}{1}
      {5}{{{\color{numbercolor}5}}}{1}
      {6}{{{\color{numbercolor}6}}}{1}
      {7}{{{\color{numbercolor}7}}}{1}
      {8}{{{\color{numbercolor}8}}}{1}
      {9}{{{\color{numbercolor}9}}}{1}
      {:}{{{\color{black}{:}}}}{1}
      {,}{{{\color{black}{,}}}}{1}
      {[}{{{\color{black}{[}}}}{1}
      {]}{{{\color{black}{]}}}}{1},
    morestring=[b]",
}

\usepackage{hyperref}
\usepackage{url}

\title{SkyScript-100M: 1,000,000,000 pairs of scripts and shooting scripts for Short Drama}

\author{Jing Tang\thanks{This work was done during an internship at Skywork AI \quad \dag{Corresponding author}}, Quanlu Jia, Yuqiang Xie\dag, Zeyu Gong\dag, Xiang Wen\\
\textbf{Jiayi Zhang, Yalong Guo, Guibin Chen, Jiangping Yang} \\
SkyWork AI \&\& Huazhong University of Science and Technology\\
\texttt{\{j\_tang,gongzeyu\}@hust.edu.cn} \\
\texttt{wenxiang@zju.edu.cn} \\
\texttt{\{luke.l,yuqiang.xie,jiayi.zhang\}@kunlun-inc.com} \\
\texttt{\{yalong.guo,guibin.chen,jiangping.yang\}@kunlun-inc.com} \\
}

\iclrfinalcopy 
\begin{document}

\maketitle

\begin{abstract}
\vspace{-5mm}
\begin{figure}[h]
\centering
\includegraphics[scale=0.43]{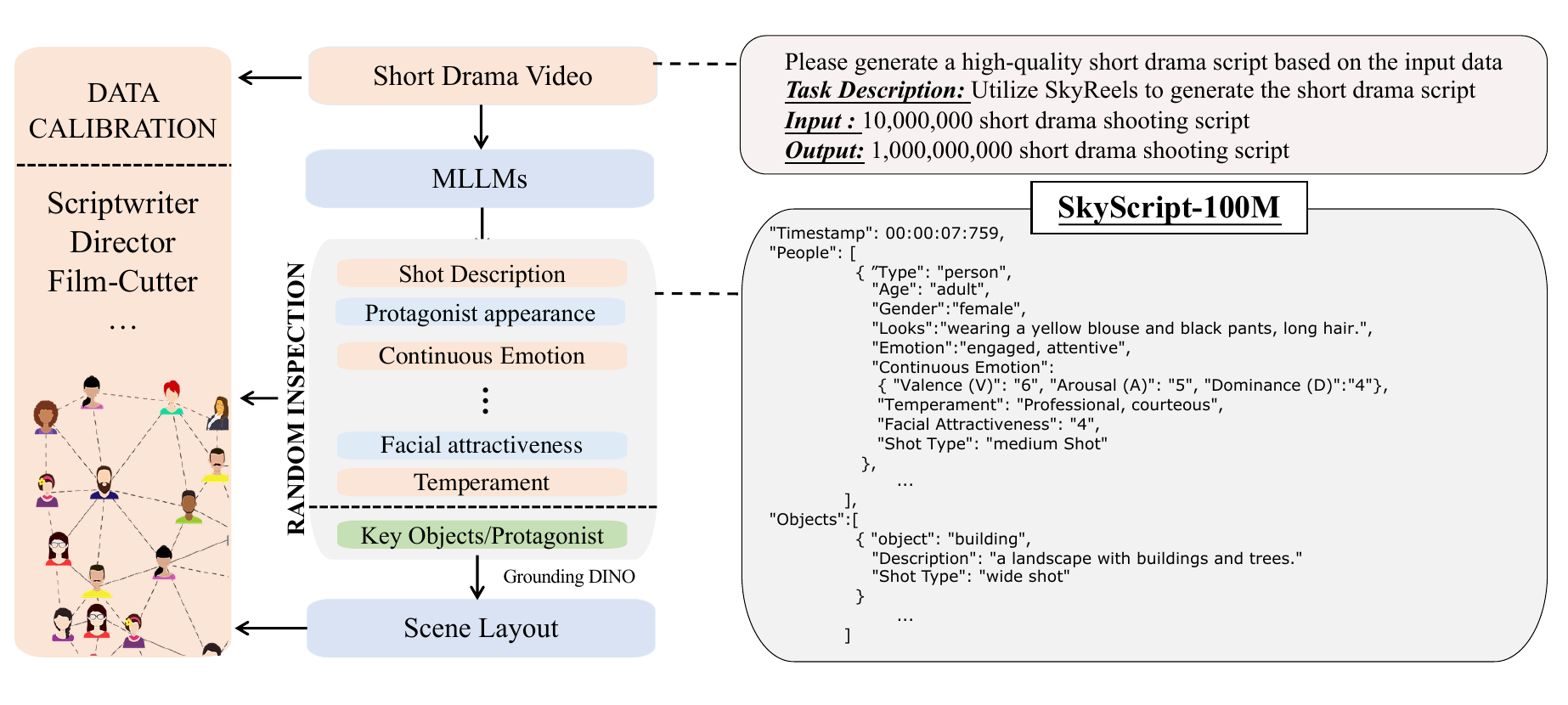}
\caption{Overview of SkyScript-100M}
\label{figure}
\end{figure}

Generating high-quality shooting scripts containing information such as scene and shot language is essential for short drama script generation. We collect 6,660 popular short drama episodes from the Internet, each with an average of 100 short episodes, and the total number of short episodes is about 80,000, with a total duration of about 2,000 hours and totaling 10 terabytes (TB). We perform keyframe extraction and annotation on each episode to obtain about 10,000,000 shooting scripts. We perform 100 script restorations on the extracted shooting scripts based on our self-developed large short drama generation model SkyReels. This leads to a dataset containing 1,000,000,000 pairs of scripts and shooting scripts for short dramas, called SkyScript-100M. We compare SkyScript-100M with the existing dataset in detail and demonstrate some deeper insights that can be achieved based on SkyScript-100M. Based on SkyScript-100M, researchers can achieve several deeper and more far-reaching script optimization goals, which may drive a paradigm shift in the entire field of text-to-video and significantly advance the field of short drama video generation.
\end{abstract}

\section{Introduction}


With the introduction of Sora \citep{openai_sora}, video generation has become one of the current popular research directions in academia \citep{opensora, Hong2023CogVideo, yang2024cogvideox} and industry\citep{gen2_runwayml, pika_art}. At the same time, the notion that AI is reshaping the film and television industry is becoming increasingly prevalent. Recently, there have been a huge number of companies conducting a series of research targeting AI-powered drama production\citep{dreamina2024, vidu2024, vimi2024}.

\begin{figure}[h]
\centering
\includegraphics[scale=0.35]{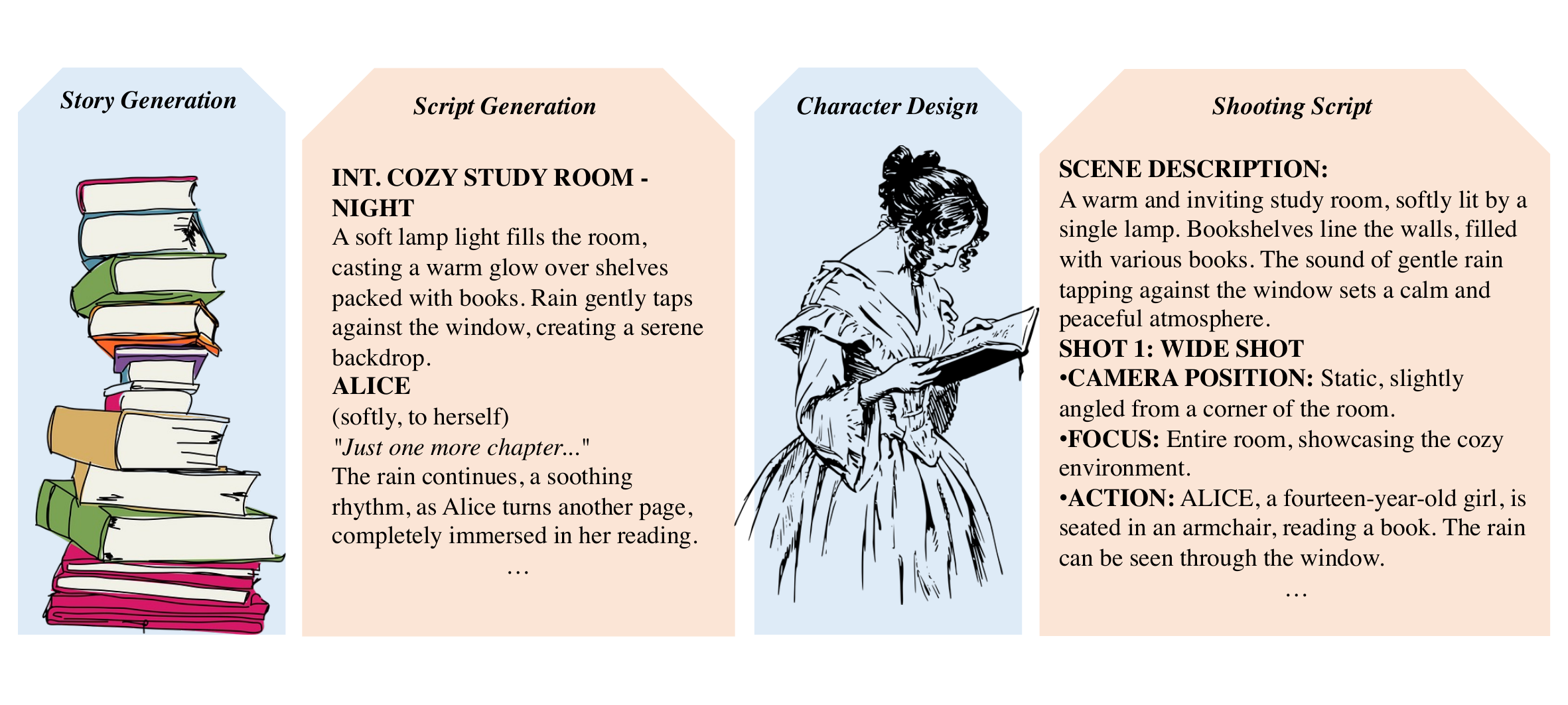}
\caption{Traditional Pipeline for Drama Production}
\label{figure-pipeline}
\end{figure}


Considering the current boom of short drama\citep{reelshort2024, shortstv2024, shortmax2024}, we focus on the task of short drama production. More specifically, we conducted in-depth discussions with professional short drama writers to uncover the real pain points in the field of short drama production, and finally focused on the generation of shooting scripts in the whole short drama production process. We analyze the traditional pipeline for drama production and explore whether such a paradigm is still suitable for AI-powered short drama production. First of all, we briefly introduce the traditional pipeline, which is shown in Figure \ref{figure-pipeline}. In this process, we begin by generating the overall story, followed by crafting the initial script based on that narrative. Once the script is completed, we proceed to design the characters, ensuring their consistency throughout. Finally, we developed the shooting script using all the preceding materials. Throughout this process, \textbf{manual adjustments to character design and key objects layout} is often necessary, which can limit the speed of the final production for short drama.

Additionally, the existing shooting script lacks annotations for key elements such as the dramatic highlights in the short drama, as well as essential promotional information like character pairing compatibility calculation for "shipping" a couple. This makes it challenging to achieve fully automated, AI-powered drama production. To address such script-to-shooting script problems while enabling truly fully automated, AI-powered drama production, we constructed a dataset containing 1,000,000,000 pairs of script and shooting scripts. More specifically, we collected 6,660 short drama episodes from the Internet, containing a total of about 80,000 episodic short dramas with a total duration of over 2,000 hours. Besides, we redefine the data structure of shooting scripts to be more suitable for the era of AI-powered drama production, which is shown in Table \ref{tab1}.


\begin{table}[htbp]
\caption{Detailed data structure of Refined shooting scripts}
\label{tab1}
\begin{center}
\begin{tabular}{|p{14cm}|}  
\hline
\multicolumn{1}{|c|}{\bf Refined shooting scripts} \\
\hline
SCENE 1 EXT. PARK - DAY \\
\hline
DESCRIPTION: \\
A sunny afternoon in a peaceful park. Birds are chirping, and a gentle breeze rustles the leaves. The park is dotted with blooming flowers and tall trees. A cobblestone path winds through the greenery. \\
\hline
SHOT 1: WIDE SHOT \\

CAMERA POSITION: Stationary, capturing the entire park scene. \\
FOCUS: A couple, EMMA (late 20s) and JACK (early 30s), walking hand in hand along the path. \\
ACTION: They walk leisurely, talking and smiling. \\
\hline
\textcolor{blue}{Highlight Score: 4.0} \\
\textcolor{blue}{3D Location of "Emma": (0,165,0)}\\
\textcolor{blue}{3D Location of "Jack": (60,180,0)}\\
\hline
Camera Detail-Time:
4 seconds \\
Camera Detail-Shot Size:
WIDE SHOT \\
Camera Detail-Camera Angle:
Eye Level \\
Camera Detail-Main Subject:
(Emma, Jack) \\
\hline
\textcolor{blue}{Event Number:
001} \\
\hline
\textcolor{blue}{Event:
Emma and Jack walking in the park} \\
\hline
\textbf{\textcolor{blue}{(Other Detailed information of characters and key objects ...)}} \\
\hline
\end{tabular}
\end{center}
\end{table}

In fact, we believe that the world in short drama can be regarded as another dimension of the world, based on this basic idea, we add as many details as possible to the world of the short drama in the shooting script, including the layout information of the key objects, the highlights of the short drama, and the changes of the emotions of the characters, so as to facilitate the LLMs to have a better view of the world in the short drama. At the same time, such annotation also allows LLMs to better understand the 3D layout of the short drama world, and ultimately achieve a more accurate screen layout. Based on this basic idea and the keyframe extraction and annotation for short drama, we completed the construction of the shooting script part of Skyscript-100M. After that, we generated new scripts based on the high-quality and structured shooting scripts and our self-developed script model, and \textbf{finally obtained 1,000,000,000 pairs of scripts and shooting scripts}.

\section{Related Works}
\subsection{Multimodal Datasets}


In order to build high-performance multimodal large language models, a large amount of high-quality visual-textual data is necessary for multimodal training tasks. Most of works \citep{xu2016msr, hendricks2017didemo, krishna2017dense,bain2021frozen} acquired video data in several different domains via web crawlers and annotated the acquired video data accordingly. However, due to the overly dispersed domains involved in these video data, this is relatively sparse for targeting a particular direction, which leads to a degradation in training performance. Some scholars collected and labelled data for professional domains such as movies \citep{rohrbach2017movie} and cooking \citep{zhou2018towards}, and constructed many high-performance domain models.Unfortunately, the short drama domain does not yet have such a mega dataset. We collected a large number of short drama videos and performed keyframe extraction on the videos, and finally obtained a huge amount of high-quality shooting script data for the construction of the final Skyscript-100M. We compare the video-text data in SkySciprt-100M with other video-text datasets in Table \ref{multimodal datasets}.

\noindent\begin{table}[htbp] 
	\centering
	\caption{Comparisons between SkySciprt-100M and other video-text datasets.}
 \label{multimodal datasets}
 \begin{tabular}{|p{5.2cm}|p{1.5cm}|p{0.8cm}|p{1cm}|p{0.9cm}|p{0.8cm}|p{1cm}|}
		\hline
		\hline Dataset & Caption & Videos & Clips  & Dur(h) & Res & Domain  \\ \hline\hline
		MSR-VTT\citep{xu2016msr}& Manual & 7.2K & 10K & 40 & 240P &open \\\hline
		DideMo\citep{hendricks2017didemo} & Manual & 10.5K & 27K & 87 & - &flickr \\ \hline
		LSMDC\citep{rohrbach2017movie} & Manual & 200 & 118K & 158 & 1080P &movie\\ \hline
		YouCook2\citep{zhou2018towards} & Manual & 2K & 14K & 176 & - &cooking\\ \hline
		How2 \citep{sanabria2018how2} & Manual & 13.2K & 80K & 2K & - &instruct\\ \hline
		ANet Caption \citep{krishna2017dense} & Manual & 20K & 100K & 849 & - &action\\ \hline
		VideoCC3M\citep{yan2021videocc3m} & Transfer & 6.3M & 10.3M & 17.5K & - &open \\ \hline
		WebVid10M \citep{bain2021frozen} & Alt-text & 10.7M & 10.7M & 52K & 360P &open\\ \hline
		WTS70M \citep{zellers2021merlot} & Metadata & 70M & 70M & 194K & - &action\\ \hline
		HowTo100M \citep{miech2019howto100m} & ASR & 1.2M & 136M & 134.5K & 240P &instruct\\ \hline
		HD-VILA-100M \citep{xu2023hdvila} & ASR & 3.3M & 103M & 371.5K & 720P &open\\ \hline
		YT-Temporal-180M \cite{zolfaghari2021yt-temporal-180m} & ASR & 6M & 180M & - & - &open\\ \hline
		InternVid \citep{wang2023internvid} & Generated & 7.1M & 234M  & 760.3K & 720P &open\\\hline
		Vript \citep{yang2024vript}& Generated & - & 420k  & 1.3K & 720P &open\\
		\hline\hline
		\textbf{SkySciprt-100M (Ours)} & \textbf{Manual} & 6.6k & 100M & 2k &720P &\textbf{drama}\\ 
		\hline\hline
	\end{tabular}
\end{table}%

\subsection{Script Generation}
Script generation is very important in natural language processing research, and has received attention from many scholars even before the emergence of large language models.Fan et al. \citep{fan2018hierarchical} creating a large WP dataset of 300K human-written stories paired with prompts, enabling hierarchical story generation, and Yao et al. \citep{yao2019plan} exploring the open-domain story generation problem, proposing a plan-and-write hierarchical generation framework based on ROC-Stories (ROC) sentence to achieve open-topic based story generation. Inheriting such ROC-based or WP-Based ideas, many  scholars \citep{peng2021towards,peng2023storyfier, huang2023affective, wang2023improving} conducted researches in this area. Unfortunately, WP and ROC contain some absurd story premises, which seriously affect the quality of the final plot.

The use of large language models for script generation is a new trend, as large language models can better provide world consensus, which means that absurd story premises in WP-based and ROC-based approaches are avoided. Mirowski et al. \citep{mirowski2022cowriting} propose a generative framework called dramatron that can uses existing, pre-trained large language models to generate long, coherent scripts. Zhou et al. \citep{zhou2023recurrentgpt} and Bai et al. \citep{bai2024longwriter} continue this line of thought for the long text coherence problem. The follow-up Ma et al. \citep{ma2024mops} focused on the modular story content management problem and Yang et al. \citep{yang2024seedstory} proposed a method for generating multimodal long script. All these frameworks are able to generate directly based on existing models, and can make good use of the potential common sense within LLM. But in the field of short Drama Script generation, there is actually a lot of background in overhead worlds. We have discussed with many film and television creators and screenwriters, and found that many common sense issues in the world of short drama are not consistent with the internal knowledge of LLMs, which leads to the final scripts still difficult to detach from the real world.



\section{The SkyScript-100M Datataset}

\noindent\begin{figure}[htbp]
\centering
\includegraphics[scale=0.33]{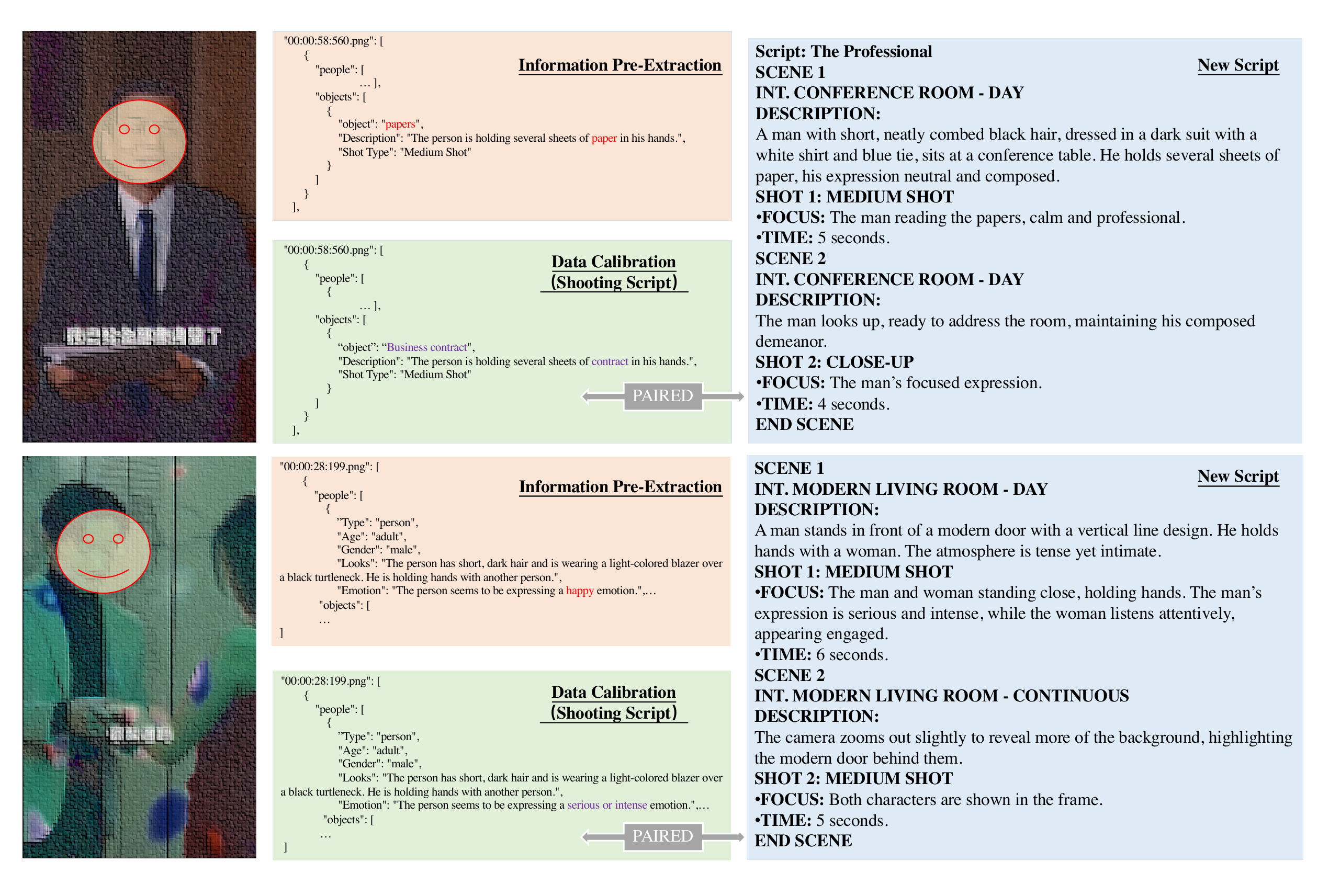}
\caption{Samples of the whole process}
\label{fig-samp}
\end{figure}

The SkyScript-100M dataset was built for the task of generating short drama videos, so we worked with professional screenwriters to develop a unique design for shooting scripts, which contain large-scale frame descriptions (including shot descriptions, character descriptions, and object descriptions) and corresponding generated scripts for short drama videos. In this section, we describe how to collect short drama videos for frame annotation and LLM-based pairwise script generation. At the same time, we will summarize and compare SkyScript-100M with existing video description datasets.
\vspace{-3mm}
\subsection{MLLMs-based pre-extraction of key information}

In order to reduce the cost of manual annotation, we first used the multimodal large language model for preliminary pre-annotation. In the specific experiments, we comprehensively compared the effects of InternVL2\citep{internvl22024}, Flash-Vstream \citep{flashvstream2024}, pLLava \citep{pllava2024} and other models, and finally chose InternVL2-Llama3-76B as the multimodal key information pre-extraction model. For the specific inference deployment, we use a server containing $32  \times$ NVIDIA A800 80G for inference acceleration. Meanwhile, in order to achieve faster pre-extraction of key information, we use the turbomind engine in lmdeploy\citep{2023lmdeploy} for further inference optimization.

It is worth noting that here, in order to achieve key information extraction that is more in alignment with the short drama world, we deeply optimize the prompt fed into our MLLMs, and ultimately use the prompt shown in Table \ref{tab2} for multimodal key information extraction. Partially examples of extracted key information of some characters and objects are shown in Table \ref{tab3} and Table \ref{tab4}.

Here, the "Continuous Emotion" in the description for the character is used to describe the current emotional state of a character. It includes three dimensions: Valence (V), which measures how positive or pleasant emotion is, ranging from negative to positive, and is used to describe the level of happiness; Arousal (A), which measures the character's level of agitation, ranging from calm or non-active to agitated or ready to act, and is used to describe the level of excitement; and Dominance (D), which measures the level of control a character feels over the situation, ranging from submissive or non-controlling to dominant or in-control, and is used to describe the sense of authority or presence. Additionally, the Temperament and Facial Attractiveness metirc are utilized in the previously mentioned character pairing compatibility calculation for "shipping" couples.

\begin{table}[hbtp]
\caption{Prompt for MLLMs-based pre-extraction of key information}
\label{tab2}
\begin{center}
\begin{tabular}{|p{14cm}|}  
\hline
\multicolumn{1}{|c|}{\bf System Prompt} \\
\hline
Your task is to analyze multiple images and identify and describe their content, focusing primarily on objects, people, and similar visual elements. \\
\hline
\multicolumn{1}{|c|}{\bf User Prompt} \\
\hline
Please perform the following steps for each image: \\
1. Foreground and Background Classification \\
\textbf{Foreground:} Identify and classify the sharper content in the image that is the primary focus. \\
\textbf{Background:} Identify and classify the blurrier content in the image that is considered non-primary content. \\
\hline
2. Shot Type Classification \\
For each identified object, use the following keywords to determine the shot type based on its position and size in the image: \\
\textbf{Wide Shot:} Captures a wide-angle view of a scene, typically showing the subject from a distance and including a lot of the surrounding background. \\
\textbf{Full Shot:} Shows a person's entire body, from head to toe, conveying the character's pose, movements, and interaction with the environment. \\
\textbf{Medium Shot:} Shoots the subject from the waist up. \\
\textbf{Close-up:} Tightly framed, focusing on a person's face or specific details of an object. \\
\textbf{Extreme Close-up:} Closer than a close-up, capturing details of the subject, such as eyes, hands, or small objects. \\
\hline
\textbf{Important Tips} \\
- Describing people uses background information or potentially occluded areas to assume a person exists. \\
- The key is to identify and describe all the people in the image! \\
- Judge the Emotion, Continuous Emotion, Temperament, and Facial Attractiveness sections, and ask everyone not to repeat the content. \\
- Use background information or potentially obscured areas to hypothesize the presence of a person when describing them. \\
- Do not repeatedly identify the same person or object. \\
- Do not identify subtitles, text, and similar elements. \\
\textbf{Output format} \\
Strictly output the description and lens type of each object in JSON format. \\
\hline
\end{tabular}
\end{center}
\end{table}

\begin{table}[hbtp]
\caption{Sample Description for Key Objects in One Frame}
\label{tab3}
\begin{lstlisting}[language=json]
"Type": "chair",
"Description": "Wooden chair with a woven backrest",
"Shot Type": "Full Shot"
\end{lstlisting}

\begin{lstlisting}[language=json]
"Type": "table",
"Description": "Round wooden table in the foreground",
"Shot Type": "Medium Shot"
\end{lstlisting}

\begin{lstlisting}[language=json]
"Type": "cup",
"Description": "Transparent glass cup filled with water",
"Shot Type": "Close-Up"
\end{lstlisting}
\end{table}

\begin{table}[hbtp]
\caption{Sample Description for Character in One Frame}
\label{tab4}
\begin{lstlisting}[language=json]
"Type": "person",
"Age": "adult",
"Gender": "male",
"Looks": "Wearing a dark suit and tie, sitting with legs crossed, has short dark hair",
"Emotion": "serious",
"Continuous Emotion": {
    "Valence (V)": "4",
    "Arousal (A)": "6",
    "Dominance (D)": "7"
},
"Temperament": "confident",
"Facial Attractiveness": "3",
"Shot Type": "Full Shot"
\end{lstlisting}
\begin{lstlisting}[language=json]
"Type": "person",
"Age": "adult",
"Gender": "male",
"Looks": "Wearing a dark suit, sitting with legs crossed, has short dark hair",
"Emotion": "neutral",
"Continuous Emotion": {
    "Valence (V)": "5",
    "Arousal (A)": "4",
    "Dominance (D)": "5"
},
"Temperament": "calm",
"Facial Attractiveness": "3",
"Shot Type": "Medium Shot"
\end{lstlisting}
\end{table}
\vspace{-3mm}
\subsection{key information cleansing and pixelation}

After completing the basic key information pre-extraction, we clean the key information and convert it into standard JSON format, which is convenient for more detailed post-processing including key object 2D detection, character 2D detection, and protagonist information post-processing. 

In this session, we extracted 500 short dramas for bad case analysis and found that they contain 134,093 valid keyframes and 5,441 invalid keyframes due to motion blur and occlusion, accounting for about 4\%. In addition, due to the format problem, the pre-extracted key information could not be read or did not meet the JSON format output requirements accounting for about 4\% of the overall data. For the JSON annotations with format problems, we use the GPT-4o \citep{openai2023gpt4} model based on the prompt shown in Table \ref{tab5} for post-processing and convert them into JSON format data that can be read directly according to the requirements. Overall, the loss rate of the pre-extracted data is 4\%.

\begin{figure}[htbp]
\centering
\includegraphics[scale=0.60]{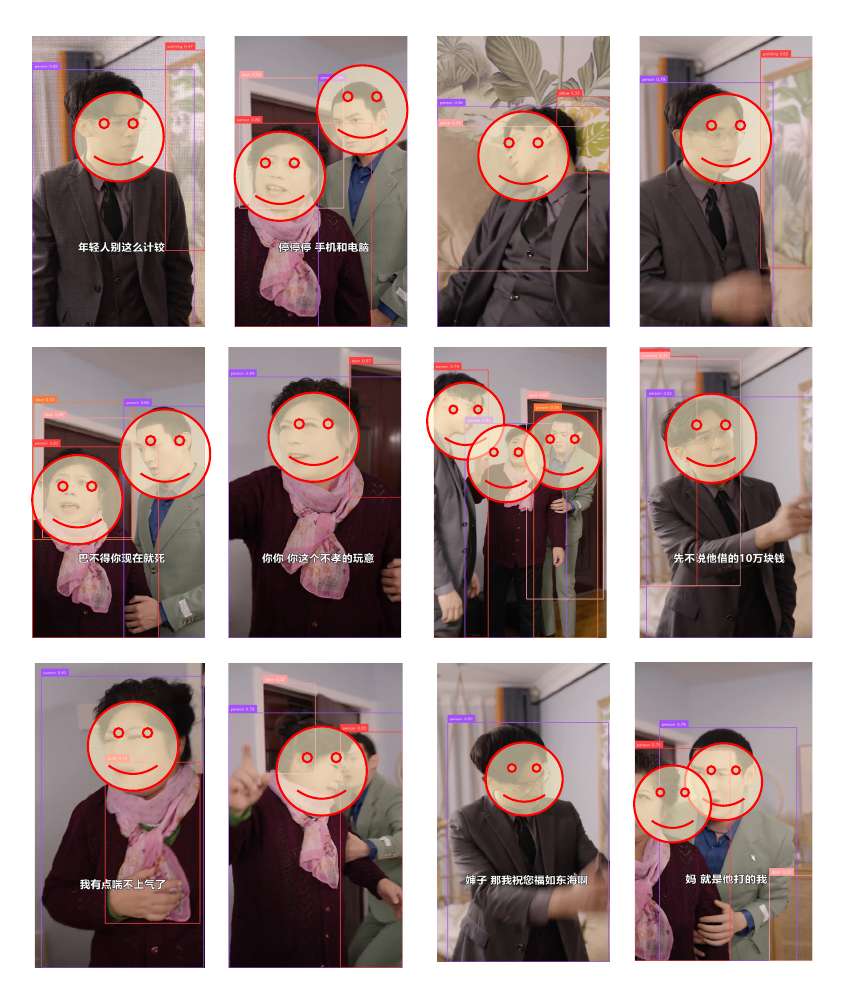}
\caption{Pixelated Facial Images for Personal Privacy Protection}
\label{fig-demo}
\end{figure}

\begin{table}[hbtp]
\caption{Prompt for JSON Correction}
\label{tab5}
\begin{center}
\begin{tabular}{|p{14cm}|}  
\hline
\multicolumn{1}{|c|}{\bf System Prompt} \\
\hline
You are a JSON validation and correction assistant. Your task is to review and correct JSON data to ensure it is properly formatted and adheres to JSON standards. Pay attention to common issues such as missing quotes, incorrect value types, misplaced commas, or brackets. \\
\hline
\multicolumn{1}{|c|}{\bf User Prompt} \\
\hline
Please review the following JSON data and correct any formatting errors. Ensure that all keys are properly quoted, all values are correctly formatted, and the overall structure is valid according to JSON standards. Provide the corrected JSON output. \\
\hline
\end{tabular}
\end{center}
\end{table}
It's worth noting that during the data cleaning process, we perform pixelation processing on face information and character name information to fully ensure ethical safety, which is shown in Figure \ref{fig-demo}. Afterward, we will mosaic the processed image and input the subsequent post-processing model of the JSON data obtained after cleaning in order to carry out further information calibration and more detailed information extraction.

\noindent
\vspace{-5mm}
\subsection{Open-vocabulary Detection using key information}

Considering the phantom problem of the multimodal large language model, we perform in-frame key object extraction based on the aforementioned cleaned JSON files. More specifically, we form an open-vocabulary list of object names identified by the multimodal large language within a single frame and the keyword $person$, and input this list into the open-vocabulary detection model along with the pixelated images for further calibration of the object positions. Here we tested the effect of different Open-vocabulary Detection models such as Grounding-DINO 1.0 \citep{liu2023grounding}, Grounding-DINO 1.5 \citep{ren2024grounding}, YOLO-World \citep{yoloworld2024}, OWLv2 \citep{minderer2023owlv2}, etc., and finally chose Grounding-DINO 1.0 as the final detection model. Some detection results are also presented in Figure \ref{fig-demo}.

\subsection{Post-processing of protagonist information}

Similarly, in order to further refine and calibrate the character information, here we first use the Deepface framework \citep{serengil2020lightface,serengil2021lightface, serengil2024lightface} to detect the face and annotate its position in the whole frame, which is used in the subsequent model for the understanding of the 2D position of the character. In addition, we also use Deepface for age, gender, emotion and ethnicity prediction of the protagonist, so as to make the information of the character more complete. Considering that there is a lot of conflicting information in the key information pre-extracted by the multimodal large language model, we refer to the Test Time Augmentation (TTA) method \citep{shanmugam2020tta} in the image detection session. More specifically, when the pre-extracted information conflicts with the results of Deepface frame ah-set detection, TTA is performed on the models called in the Deepface framework, and the detection results of the Deepface framework are finally adopted.

In addition to using Deepface to correct the key information of the face, we also use DepthAnything v2 \citep{depthanythingv22024} to correct the depth of field of the character, and use AlphaPose \citep{alphapose, fang2017rmpe, li2019crowdpose} to discriminate the posture of the character. Finally, we have achieved a richer and more accurate character information construction.

\subsection{Data Calibration}
Here we set up a data calibration team consisting of 12 annotators who have deep communication with short drama scriptwriters and are fully aligned with the evaluation criteria. The whole calibration process includes replenishment of omitted information, correction of mislabelled information, and removal of redundant annotations, etc. We sampled the final results in the form of data packets. The final labeling results were randomly sampled in the form of data packets, ranging from 200-500 random data packets, and the accuracy rate of the sampling was not up to standard, and then returned to repair, and the final accuracy rate exceeded 90\%, which was in line with the short drama production requirements.

\section{New Short Drama Generation Paradigm}

The traditional pipeline for drama generation is shown in Table \ref{table-pipe}, which consists of Story Generation, Script Generation, Character Design and finally Shooting Script Generation. The most important part is Shooting Script Generation, which determines the quality of the final generated drama video. 

\begin{figure}[htbp]
\centering
\includegraphics[scale=0.55]{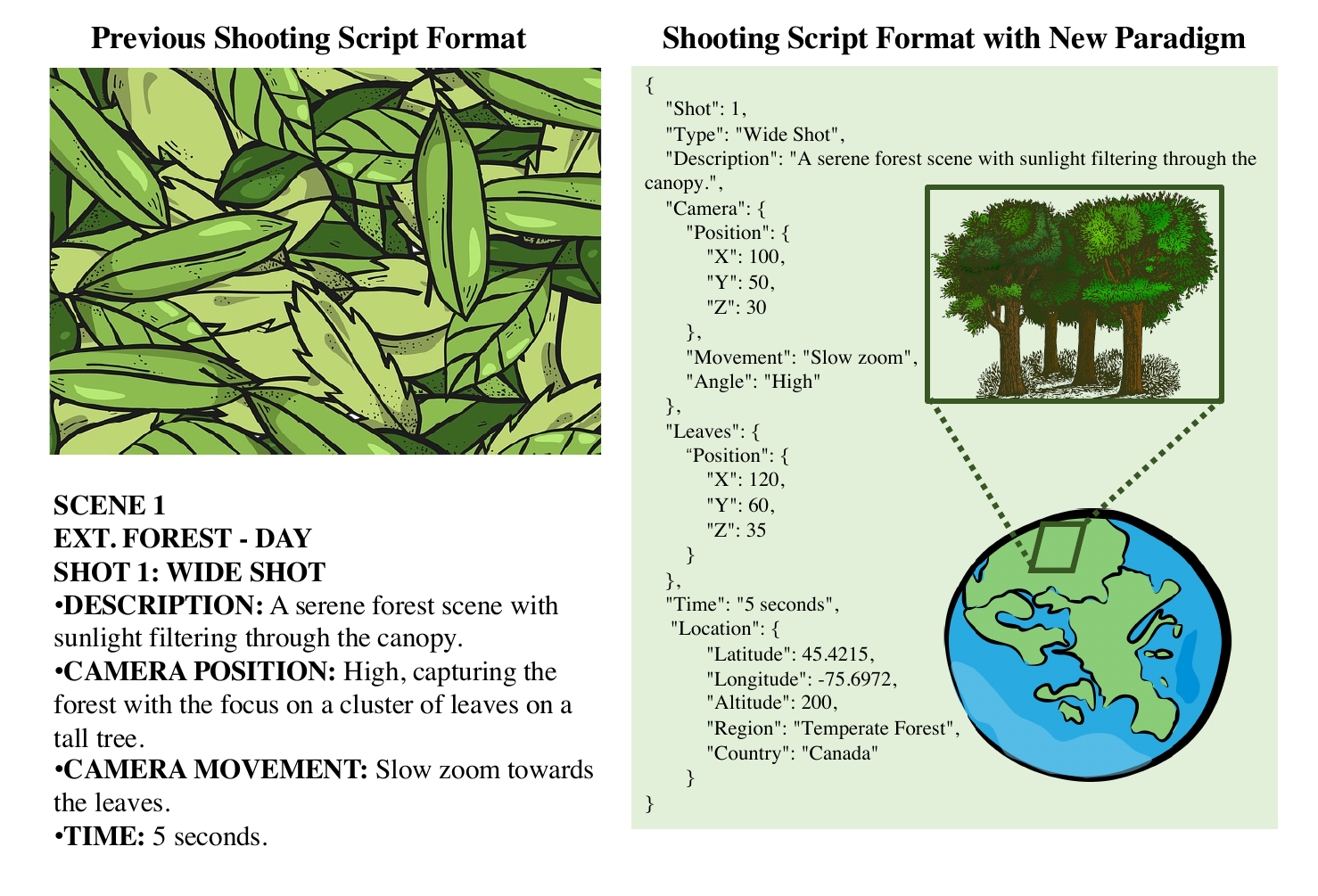}
\caption{Previous Shooting Script Format and Shooting Script Format with New Paradigm}
\label{fig-new}
\end{figure}

As shown in Figure \ref{fig-new}, the Shooting Script usually consists of basic parameters such as scene description, camera position, camera movement, camera movement time, etc.  However, it is difficult for the video generation model to understand such information directly, and a more powerful video generation method is feed as much information as possible from the short drama world to video generation model, so that it has the same world consensus as the short drama creators in the generation session. For this reason, we define the Shooting Script format with new paradigm, which containing as much information as possible, so that it can maintain a consistent memory of characters, objects, and even world concepts throughout the short drama world during the multi-frame generation session. We named it the process of \textbf{aligning the video generation model with short drama world}.


 \begin{table}[htbp]
 \caption{Pipeline for drama generation}
 \label{table-pipe}
 \begin{center}
 \begin{tabular}{|p{14cm}|}  
 \hline
 \multicolumn{1}{|c|}{\bf Story Generation} \\
 \hline
Elara, a young woman in a quiet village, harbored a dark secret. Every full moon, she disappeared into the forest, returning the next morning, exhausted and silent. The villagers were curious but never questioned her. Unbeknownst to them, Elara was cursed—under the full moon's light, she transformed into a werewolf, prowling the night. She bore this burden alone, protecting her village from the beast within, knowing she could never reveal the truth of her curse.\\
 \hline
 \multicolumn{1}{|c|}{\bf Script Generation} \\
 \hline
 SCENE 1 \\
 
 EXT. VILLAGE - NIGHT \\
 
 DESCRIPTION: \\
 The village sleeps under the full moon. Elara silently slips away into the forest, her expression tense. \\
 
 SHOT 1: WIDE SHOT \\
 
 CAMERA POSITION: High angle, showing Elara leaving the village. \\
 TIME: 5 seconds. \\
  \hline
 SCENE 2 \\
 
 EXT. FOREST CLEARING - NIGHT \\
 
 DESCRIPTION: \\
 In a moonlit clearing, Elara clutches her chest as the transformation begins. Her body contorts, and she becomes a werewolf. \\
 
 SHOT 2: MEDIUM SHOT \\
 
 CAMERA POSITION: Eye level, capturing the start of her transformation. \\
 TIME: 8 seconds. \\
  \hline
 \multicolumn{1}{|c|}{\bf Character Design} \\
 \hline
 \textbf{Elara}: female, mid-20s, long dark hair, quiet, brave, burdened by a secret curse. \\
 \hline
 \multicolumn{1}{|c|}{\bf Shooting Script Generation} \\
 \hline
 \multicolumn{1}{|c|}{\bf \textcolor{red}{KEY POINT}} \\
 \hline
 \end{tabular}
 \end{center}
 \end{table}

\noindent

\begin{figure}[h]
\centering
\includegraphics[scale=0.35]{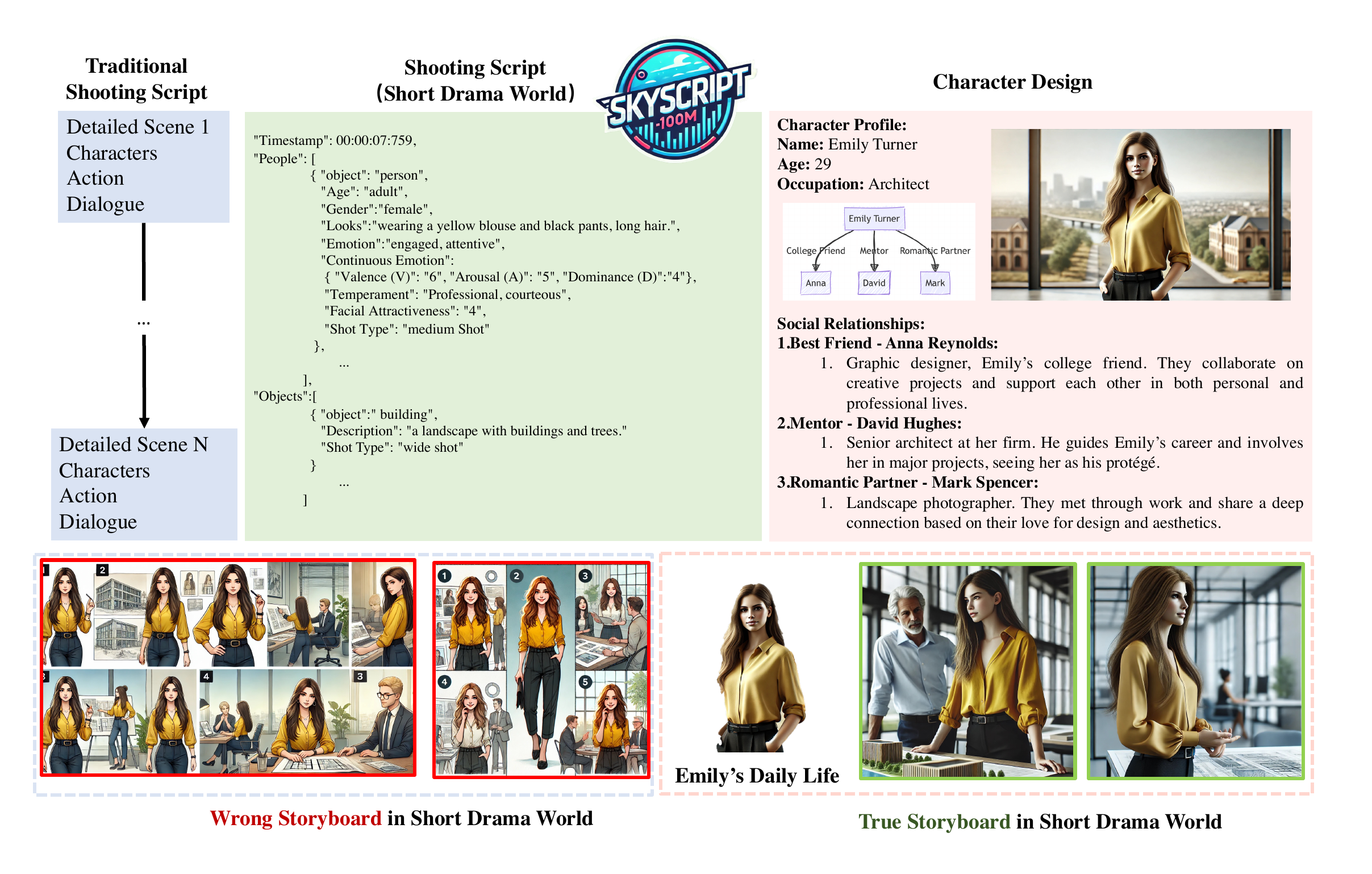}
\caption{New Short Drama Video Generation Paradigm}
\label{fig-skyreel}
\end{figure}

\begin{figure}[htbp]
\centering
\includegraphics[scale=0.5, trim =10 10 10 10,clip]{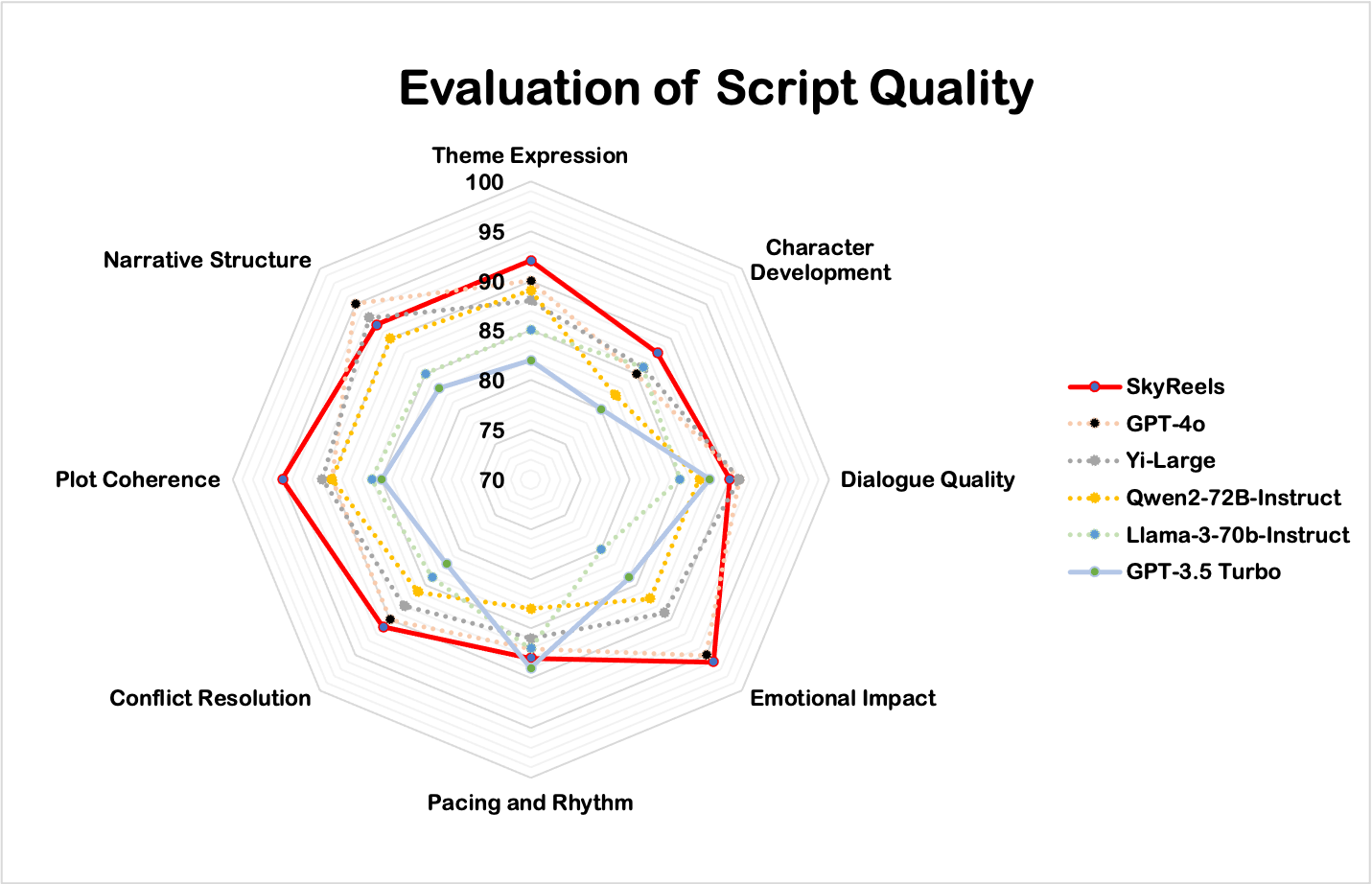}
\caption{Evaluation of Large Script Model Quality}
\label{fig-eva}
\end{figure}

We apply such a new short drama generation paradigm to our large short drama generation model SkyReels, and explore the effectiveness of this new paradigm through a large number of generation experiments. We evaluated SkyReels across various dimensions, including Theme Expression, Character Development, Dialogue Quality, Emotional Impact, Pacing and Rhythm, Conflict Resolution, Plot Coherence, and Narrative Structure. The results indicate that SkyReels demonstrates outstanding performance in these areas compare with other advances LLMs \citep{openai2024gpt4o,yi-large2024,yang2024qwen2,smith2024llama3, openai2023gpt35}. The detailed experimental results are presented in Figure \ref{fig-eva}.One specific generation details and experimental results are shown in Figure \ref{fig-skyreel}. It can be found from the experimental results that the videos generated based on shooting script under the new paradigm can better ensure the consistency of characters and painting style, and can have a deeper understanding of the relationship between characters and the trend of plot in the short drama world.

\noindent

\section{Broad Impact on Short Drama}

Based on the huge amount of paired "Script-Shooting Script" data in SkyScript-100M, it is possible to carry out very broad and in-depth research on skit generation from various perspectives. In this section, we conduct a preliminary exploratory study and optimize our large short drama generation model SkyReels, based on these findings.

\subsection{Video Highlight Detection}

With the explosive growth of short drama, automatic video highlight detection technology for short drama becomes more and more urgent. Highlight detection can extract the emotional peak and unexpected or long-planned plot in the whole story, which is convenient for editing and promotion in the post-production of a short drama. However, due to the characteristics of short dramas, it is difficult to extract the highlights based on straightforward quantitative values such as barrages and likes, because the target users of short dramas seldom send barrages, and the likes of short dramas are for whole episodes, which makes it difficult to extract the highlights of single episode. Some scholars have conducted research on video highlight detection based on biosignals \citep{7010926,katsigiannis2018dreamer, correa2018amigos}, but this type of data collection is too expensive and difficult to construct a super-large dataset for a particularly comprehensive short drama analysis. Meanwhile, popular short drama is updated fast, and it is difficult to achieve real-time updates of datasets with bioelectric signals.


In order to achieve accurate and reliable highlight annotation, we start from Plutchik emotion theory \citep{plutchik1980emotion} and analyze the emotions of the characters in short drama single episodes from the perspectives of Valence (V), Arousal (A), and Dominance (D), and finally get the video highlight score in every single episode in short drama. It is worth noting that, unlike previous single-frame discrete highlight annotations, here we set up a continuous highlight score estimator for assessing the change in short drama, which means that based on our data, regression-based highlight detection methods for short drama videos can be available.

\noindent\begin{figure}[htbp]
\centering
\includegraphics[scale=0.3]{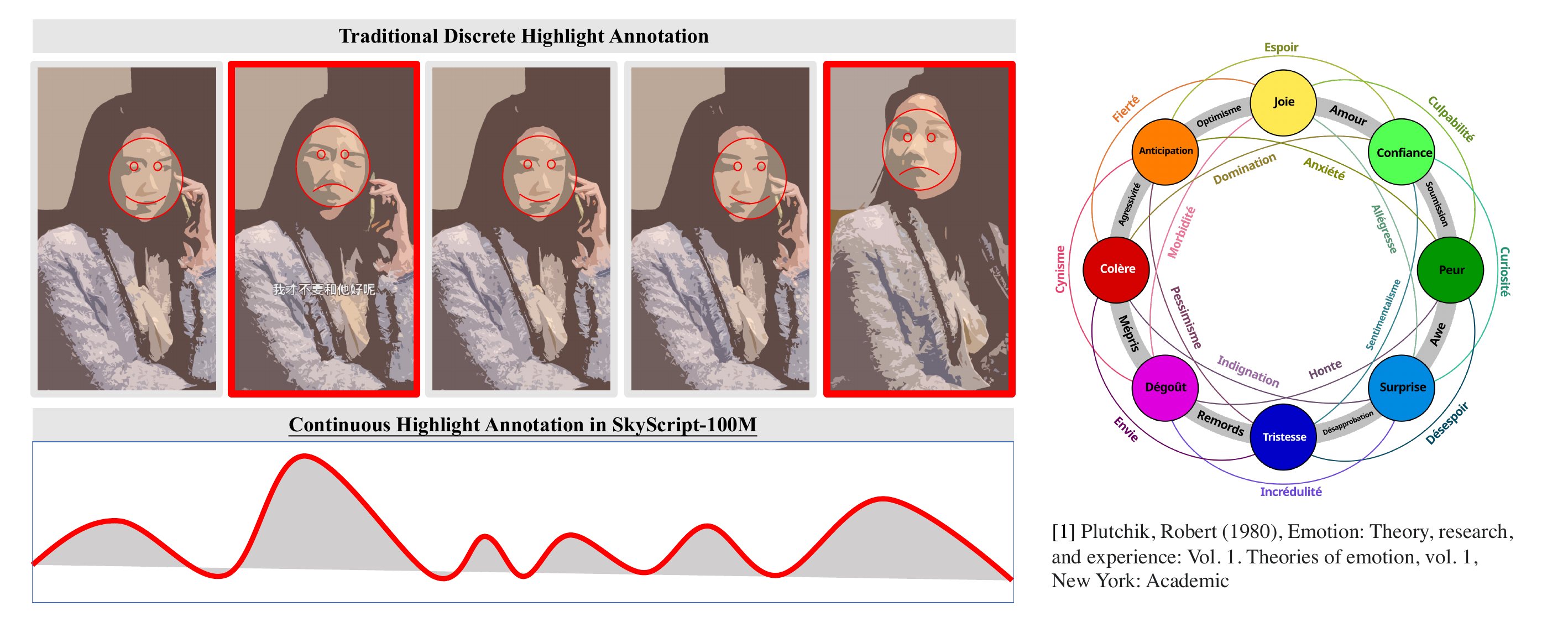}
\caption{Continuous Video Highlight Annotation in SkyScript-100M}
\label{figure}
\end{figure}

\subsection{World Layout Understanding}

World layout understanding is an indispensable part of the screen generation task, however, most of the existing short drama generation models are difficult to achieve accurate and reliable world layout understanding and character position planning, which can lead to abnormal character or object jitter or even transient movement in the final generated video. To address the world layout understanding problem, we measure and estimate the character positions in the whole short drama to obtain a rough 2D position, and then inversely solve the 3D position using the Multiple View Geometry theory and obtain the 3D-2D transformation matrix to regenerate new 2D view, which is shown in \ref{eq1}. 

\begin{equation}
\left(\begin{array}{c}
u \\
v \\
1
\end{array}\right)=\left(\begin{array}{cccc}
t_1 & t_2 & t_3 & t_4 \\
t_5 & t_6 & t_7 & t_8 \\
t_9 & t_{10} & t_{11} & t_{12}
\end{array}\right)\left(\begin{array}{c}
x \\
y \\
z \\
1
\end{array}\right)
\label{eq1}
\end{equation}

\begin{figure}[htbp]
\centering
\includegraphics[scale=0.32]{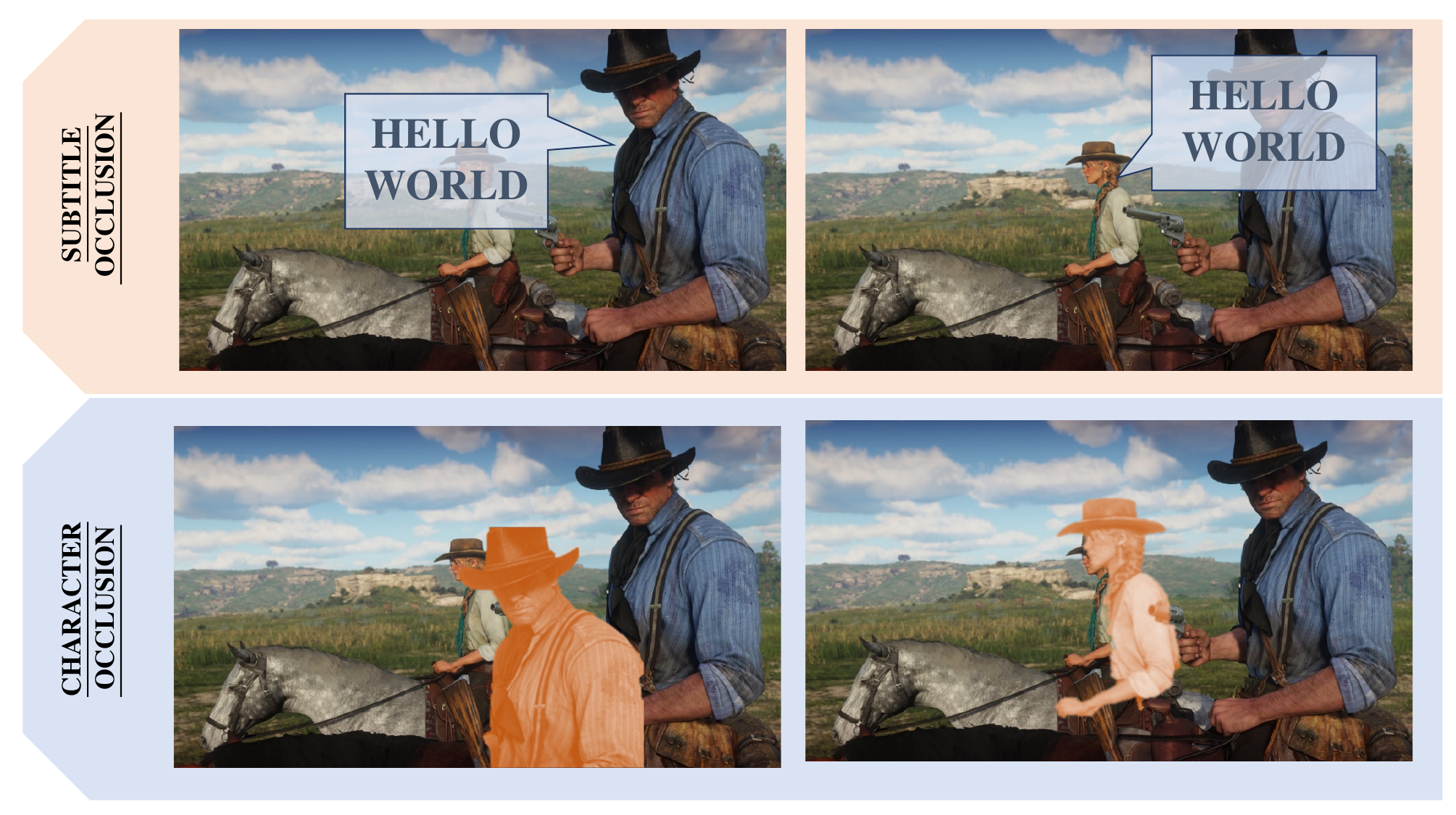}
\caption{Anomalous occlusion in 3D short drama}
\label{fig-g7}
\end{figure}

We constrain the positions of characters and objects on the screen to conform to the laws of the objective physical world through consistency constraints based on the 3D positions and the logical chain of position calculation. Meanwhile, in some 3D filming applications, the problem of anomalous occlusion as shown in Figure \ref{fig-g7} often occurs. These problems can also be circumvented based on an accurate and reliable understanding of the world layout.

\noindent\begin{figure}[htbp]
\centering
\includegraphics[scale=0.32]{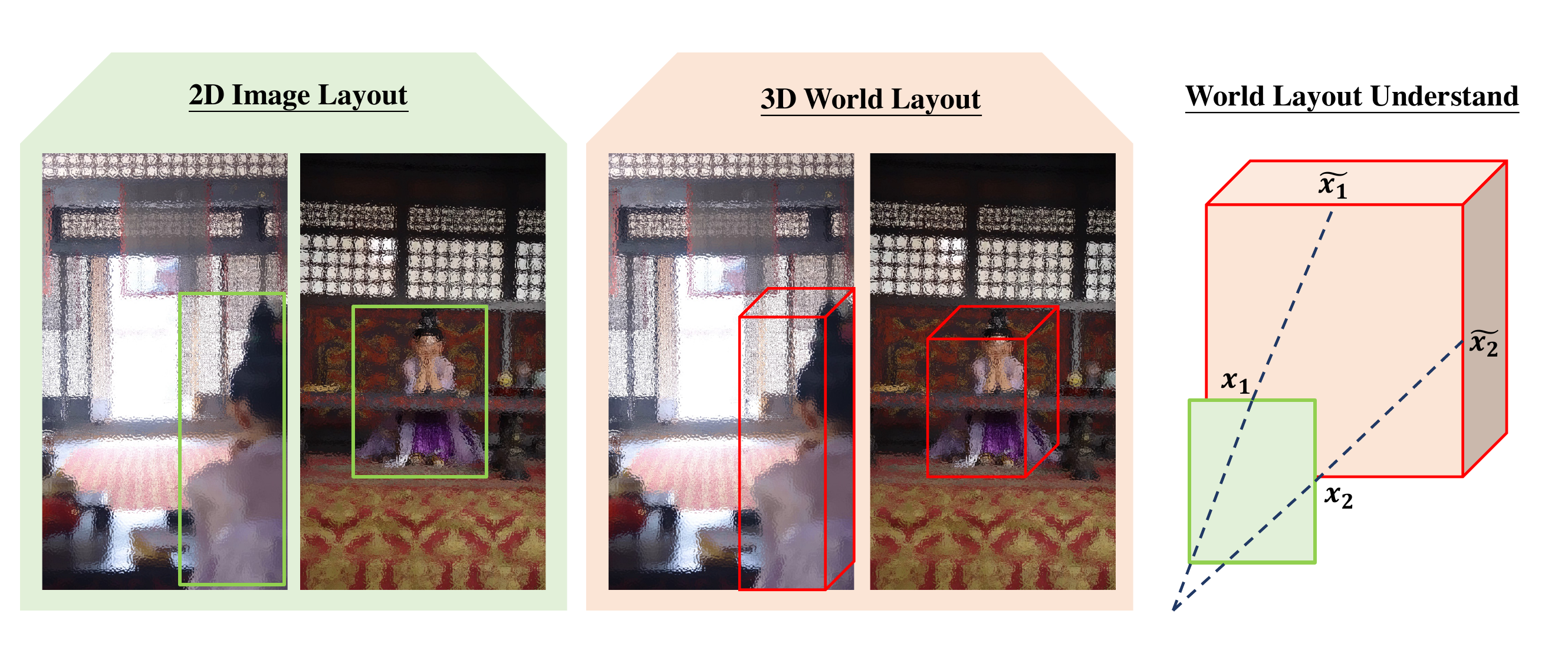}
\caption{World Layout Understand in SkyScript-100M}
\label{figure}
\end{figure}

\noindent

\noindent\begin{figure}[htbp]
	\centering
	\includegraphics[scale=0.26]{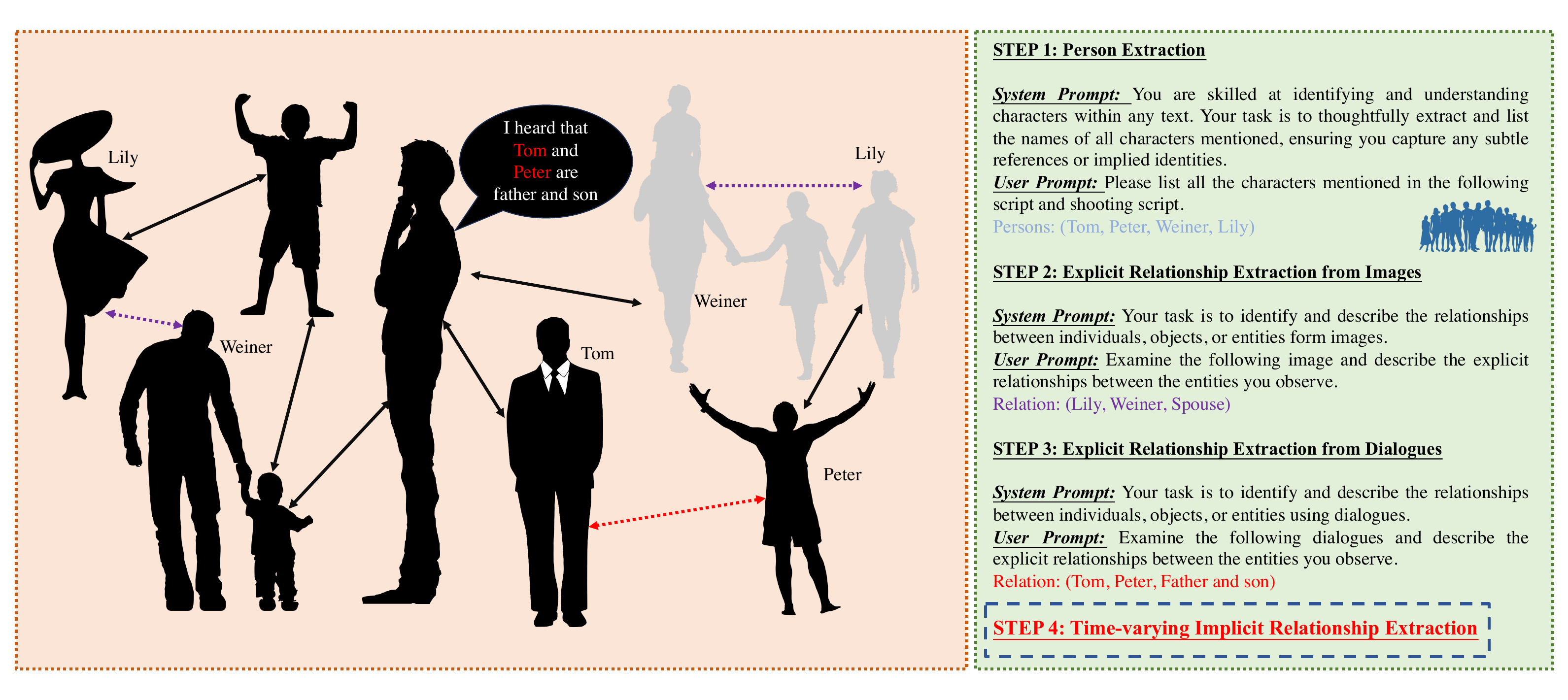}
	\caption{Implicit Character Relationship Mining}
	\label{fig-re}
\end{figure}

\noindent\begin{figure}[htbp]
	\centering
	\includegraphics[scale=0.5]{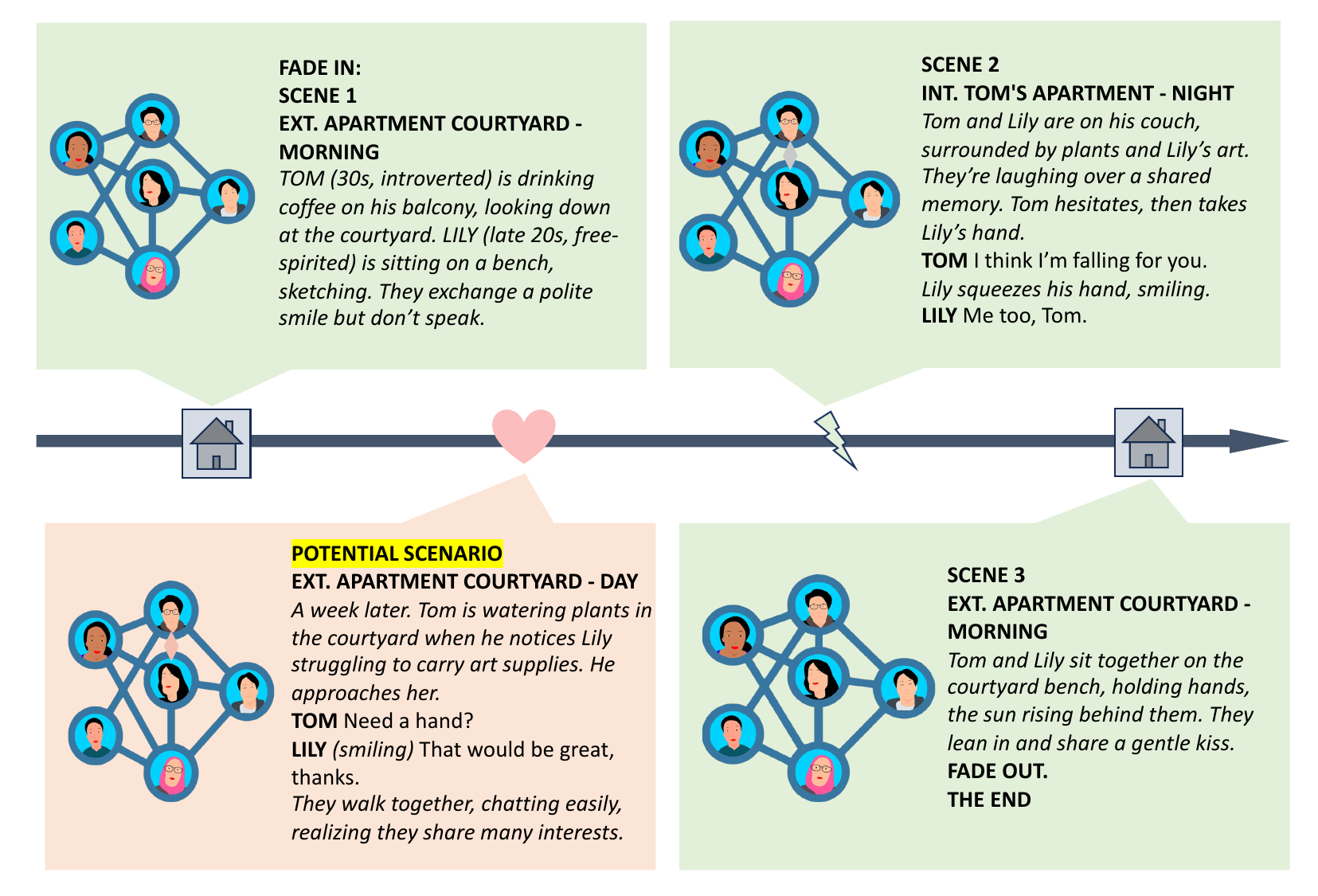}
	\caption{Counterfactual Script Generation in SkyScript-100M}
	\label{fig-cou}
\end{figure}

\subsection{Implicit Character Relationship Mining}
In traditional scripts, character relationships are usually saved in the form of very fixed explicit triples, but human emotions change all the time, and many explicit relationships implicit relationships between characters. The emergence of large language models makes time-varying implicit character relationship mining possible. We constructed a pipeline based on the paired "Script-Shooting Script" data in Skyscript-100M, as shown in Figure \ref{fig-re}. 

Firstly, we perform entity extraction on the characters in the story to construct a character library. After that, we mine the implicit relationship information from images and Skyscript-100M annotated text based on multimodal and unimodal large language models, respectively. It is worth noting that since we annotate the character location information and continuous emotions here, the text-based mining can also have high recognition accuracy. Based on the mined time-varying implicit character relationship, the script generation model can better consider the complexity and uncertainty of the plot in the real social scene in short drama world, and can better understand the subtle relationship in the narrative context, which lead to new hidden key new plots that are counter-intuitive but reasonable shown in Figure \ref{fig-cou}.

\section{Conclusion}


We pioneered the construction of a multimodal dataset for short drama, and based on this dataset we derived 1,000,000,000 pairs of scripts and shooting scripts to construct the SkyScript-100M. We describe the construction process of SkyScript-100M in detail, and conduct an in-depth study based on SkyScript-100M, proposing a new short drama generation paradigm.In addition, we explore the broad impact of SkyScript-100M on short drama, and explore its potential in advancing short drama video generation by addressing key points such as video highlight detection, world layout understanding and implicit character relationship mining. We will continue to improve the series of SkyScript and optimize our large short drama generation model SkyReels based on SkyScript.

\bibliography{iclr2023_conference}
\bibliographystyle{iclr2023_conference}


\end{document}